\documentclass[letterpaper, 10 pt, conference]{ieeeconf}
\usepackage[T1]{fontenc}
\IEEEoverridecommandlockouts                                  
\usepackage{times}
\usepackage{epsfig}
\usepackage{graphicx}
\usepackage{amsmath}
\usepackage{amssymb}
\usepackage{multirow}
\usepackage{comment}
\usepackage{color}
\usepackage{multirow}
\usepackage{makecell}
\usepackage{times}
\usepackage{bbding}
\usepackage{pifont}
\usepackage{float}
\usepackage{cite}
\usepackage{threeparttable}

\usepackage[hidelinks,colorlinks=true,linkcolor=blue,citecolor=blue,urlcolor=blue]{hyperref}
\usepackage{booktabs}
\usepackage[linesnumbered,ruled,vlined]{algorithm2e} 
\SetKwInput{KwIn}{Input}    
\SetKwInput{KwOut}{Output}
\usepackage[table]{xcolor}  
\usepackage{tabularx}       
\usepackage{colortbl}       
\usepackage{array}          
\definecolor{metricblue}{RGB}{220, 230, 250}
\definecolor{headergray}{gray}{0.9}
\usepackage{listings}
\usepackage[many]{tcolorbox}
\tcbuselibrary{listingsutf8}
\usepackage{float}
\usepackage{booktabs}
\usepackage{colortbl}

\newtcblisting{myjsonbox}[2][]{%
  float,
  floatplacement=htbp,
  listing only,
  colback=gray!5,
  colframe=black,
  boxrule=0.4pt,
  arc=2pt,
  left=5pt,right=5pt,top=5pt,bottom=5pt,
  title={#2},
  listing options={
    language=Java,
    basicstyle=\ttfamily\footnotesize,
    breaklines=true,
    showstringspaces=false,
    columns=fixed,
    keepspaces=true,
  },
  #1
}

\title{\LARGE \bf
Talk Less, Fly Lighter: Autonomous Semantic Compression for UAV Swarm Communication via LLMs}

\author{
Fei Lin$^{1}$ $^{*}$, Tengchao Zhang$^{1}$ $^{*}$, Qinghua Ni$^{1}$, Jun Huang$^{1}$, Siji Ma$^{1}$, \\Yonglin Tian$^{2}$ $^{\dagger}$, Yisheng Lv$^{2}$ $^{\dagger}$, and Naiqi Wu$^{1}$ 
\thanks{$^{*}$Fei Lin and Tengchao Zhang contribute equally to this work.}
\thanks{$^{\dagger}$Corresponding author: Yonglin Tian, Yisheng Lv (e-mail: \{yonglin.tian, yisheng.lv\}@ia.ac.cn).}
\thanks{This work was partly supported by the Science and Technology Development Fund, Macau Special Administrative Region (SAR) (0145/2023/RIA3, 0093/2023/RIA2, 0157/2024/RIA2)}
\thanks{$^{1}$Fei Lin, Tengchao Zhang, Qinghua Ni, Jun Huang, Siji Ma, and Naiqi Wu are with the Department of Engineering Science, Faculty of Innovation Engineering, Macau University of Science and Technology, Macau 999078, China (e-mail: \{feilin, zhangtengchao, qinghua.ni, junhuang, siji.ma\}@ieee.org, nqwu@must.edu.mo.)}
\thanks{$^{2}$Yonglin Tian and Yisheng Lv are with the State Key Laboratory for Management and Control of Complex Systems, Institute of Automation, Chinese Academy of Sciences, Beijing 100190, China (e-mail: \{yonglin.tian, yisheng.lv\}@ia.ac.cn).}
}

\begin{document}

\maketitle
\thispagestyle{empty}
\pagestyle{empty}


\begin{abstract}

The rapid adoption of Large Language Models (LLMs) in unmanned systems has significantly enhanced the semantic understanding and autonomous task execution capabilities of Unmanned Aerial Vehicle (UAV) swarms. However, limited communication bandwidth and the need for high-frequency interactions pose severe challenges to semantic information transmission within the swarm. This paper explores the feasibility of LLM-driven UAV swarms for autonomous semantic compression communication, aiming to reduce communication load while preserving critical task semantics. To this end, we construct four types of 2D simulation scenarios with different levels of environmental complexity and design a communication-execution pipeline that integrates system prompts with task instruction prompts. On this basis, we systematically evaluate the semantic compression performance of nine mainstream LLMs in different scenarios and analyze their adaptability and stability through ablation studies on environmental complexity and swarm size. Experimental results demonstrate that LLM-based UAV swarms have the potential to achieve efficient collaborative communication under bandwidth-constrained and multi-hop link conditions.

\end{abstract}


\section{Introduction}

Thanks to their efficiency and flexibility in tasks such as collaborative monitoring, disaster rescue, and environmental modeling, UAV swarms have become a key focus in both industry and academia~\cite{javed2024state}. Meanwhile, Foundation Models (FMs) such as Large Language Models (LLMs) and Multimodal Large Language Models (MLLMs) have made significant progress in semantic reasoning, cross-modal perception, and spatial relationship understanding~\cite{zhu2024chatnav, luo2025visual, zhou2024learning, gong2025space}, and have demonstrated outstanding task comprehension and decision-making capabilities in fields such as autonomous driving and autonomous robotics~\cite{zhou2025towards, yue2024deer}. This technological trend is gradually extending to UAV systems, offering new possibilities for achieving higher levels of intelligence in scenarios such as mission planning, environmental perception, and multi-UAV collaboration~\cite{tian2025uavs, wang2024tasf, lin2025airvista, zhang2025logisticsvln}.

In FM-driven UAV swarms, complex semantic information must be frequently exchanged among multiple UAVs to support dynamic task allocation and collaborative decision-making~\cite{zhang2025coordfield}. Traditional centralized or distributed architectures rely on structured communication protocols such as MAVLink~\cite{campion2018uav}, which are suitable for low-level control and fixed tasks but struggle to carry high-level semantic information that includes context and task intent. Introducing natural language communication can fully exploit the reasoning capabilities of FMs. Still, under low-bandwidth and multi-hop forwarding conditions, long instructions are prone to expansion or repetition during progressive transmission, leading to latency accumulation and redundant propagation, thereby affecting collaboration efficiency~\cite{ping2025multimodal}.

To address the above issues, a natural approach is to leverage the semantic modeling capabilities of FMs to achieve autonomous semantic compression in communication, thereby reducing load while preserving key information. Although existing studies have explored the capabilities of LLMs in instruction simplification, context compression, and semantic fidelity evaluation~\cite{jiang2023longllmlingua, liskavets2025prompt}, systematic methods and empirical evaluations for low-bandwidth, multi-hop UAV swarms remain lacking. To this end, this paper presents an exploratory study that employs an LLM-driven autonomous semantic compression mechanism, verifying the feasibility and potential of LLM-based autonomous UAV swarms in semantic compression communication. Specifically, the main contributions of this paper are as follows:

\begin{itemize}
    \item We propose a pipeline for LLM-driven UAV swarms that supports autonomous semantic compression, communication, and task execution.
    \item We construct four types of 2D simulation environments with different levels of complexity and integrate a multidimensional evaluation framework to comprehensively assess the trade-off between semantic integrity and communication efficiency.
    \item We evaluate the semantic compression performance of nine mainstream LLMs and, through ablation studies, reveal the effects of environmental complexity and swarm size on semantic compression.
\end{itemize}

\section{Preliminaries}

\subsection{FMs for UAV Swarm Communication}

The introduction of FMs provides new optimization approaches for high-level task information transmission in UAV swarms~\cite{jiang2025large, kaleem2024emerging}. MRLMN~\cite{xu2025scalable} leverages LLMs to assist multi-agent reinforcement learning, enabling multi-hop network planning and thereby improving the communication coverage and transmission efficiency of UAV swarms. Zhang \emph{et al.}~\cite{yan2025hierarchical} deploy LLMs on both High-altitude Platform Stations (HAPS) and UAV terminals, and by jointly optimizing communication access strategies, bandwidth allocation, and flight path control, they effectively enhance swarm link quality and task execution performance. Notably, RAMSemCom~\cite{liu2025wireless} draws on the idea of Retrieval-Augmented Generation (RAG), enabling FM-based multi-agent systems to achieve efficient multimodal semantic communication under bandwidth-constrained conditions; in principle, this framework can be transferred to UAV swarm scenarios. Although the above studies demonstrate the potential of FMs in optimizing swarm communication, for UAV swarm communication scenarios where low bandwidth, multi-hop forwarding, and task semantic propagation requirements coexist, systematic methods and empirical validation for semantic compression remain lacking. 

\subsection{FMs for Semantic Compression}

In sequence modeling, FMs rely on autoregressive prediction and context dependency capture, enabling them to implicitly model redundant information while preserving core semantics during generation, thereby possessing an inherent capability for semantic compression. Early studies primarily focused on textual scenarios, where LLMs were employed to compress long texts or prompt instructions, reducing transmission overhead while maintaining semantic consistency~\cite{gilbert2023semantic}. In recent years, this capability has been introduced into the field of semantic communications. For example, KG-LLM~\cite{salehi2025llm} combines knowledge graphs with LLMs to achieve efficient transmission of structured semantics, with related works including LLM-SC~\cite{wang2025large} and others. Notably, MLLMs can further assess task relevance for multimodal data, prioritizing the compression of key information to reduce transmission load while ensuring semantic fidelity, with related research including MLLM-SC~\cite{zhang2025multimodal}, M4SC~\cite{jiang2025m4sc}, and others.


\section{Approach}

\begin{figure}[htbp]
    \centering
    \includegraphics[width=0.95\linewidth]{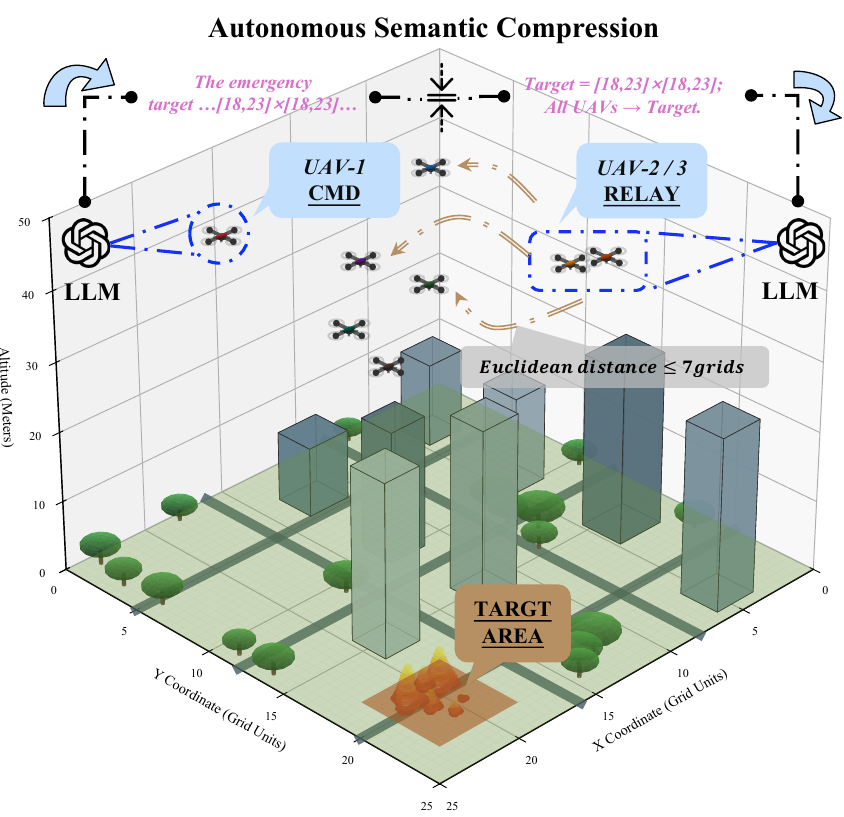}
    \caption{3D illustration of the LLM-driven communication compression workflow in the Extreme scenario. The actual simulation is 2D, with 3D rendering used for intuitive visualization of UAV swarm coordination and task flow.}
    \label{fig:1}
\end{figure}

To verify whether an LLM-based UAV swarm can achieve efficient communication through autonomous semantic compression, we constructed four types of 2D simulation environments with varying complexity levels, and designed corresponding system prompts and task instructions based on the characteristics of the scenarios and UAV behavior attributes. Fig.~\ref{fig:1} illustrates an example process of the proposed method, in which the LLM is used as a ``communication compression engine'' that, guided by prompts, performs compression and translation of natural language instructions, retaining key semantics to the greatest extent and enabling efficient inter-swarm communication and collaborative task execution.

\subsection{Simulation Environments and Pipeline}

We divide the 2D simulation test environments into four categories: simple, standard, complex, and extreme scenarios, with key parameters shown in Table~\ref{tab:uav_scenarios}. Due to the limited number of UAVs, the first three scenarios all adopt relatively flat communication structures: in the simple scenario, the two UAVs communicate directly via point-to-point links without the need for relay nodes; in the standard scenario, the three UAVs form a triangular communication network, with one UAV serving as a commander node capable of communicating with all others; the complex scenario adopts a star topology, where UAV-1 acts as the central node directly connected to the other four UAVs, thus avoiding the complexity introduced by multi-hop relays. Due to space constraints, this paper focuses on the extreme scenario as the primary subject of analysis.

\begin{table*}[htbp]
  \small
  \renewcommand{\arraystretch}{1.2}
  \caption{Key parameters of the four 2D simulation scenarios for UAV swarm communication.}
  \label{tab:uav_scenarios}
  \centering
  \begin{tabular*}{\textwidth}{@{\extracolsep{\fill}}l|c|c|c|c|c|c}
    \toprule
    \textbf{Scenario Name} & \textbf{Number of UAVs} & \textbf{Env. Size} & \textbf{Target Area} & \textbf{Comm. Range} & \textbf{Max Time (s)} & \textbf{Max Steps} \\
    \midrule
    Simple     & 2 & 10×10 & [8,9]×[8,9]     & Full Range & 75  & 15  \\
    Standard   & 3 & 15×15 & [12,14]×[12,14] & Full Range & 125 & 25  \\
    Complex    & 5 & 20×20 & [16,19]×[16,19] & Full Range & 160 & 32  \\
    Extreme    & 8 & 25×25 & [18,23]×[18,23] & 7.0 grids  & 200 & 40  \\
    \bottomrule
  \end{tabular*}
\end{table*}

\begin{figure}[htbp]
    \centering
    \includegraphics[width=0.95\linewidth]{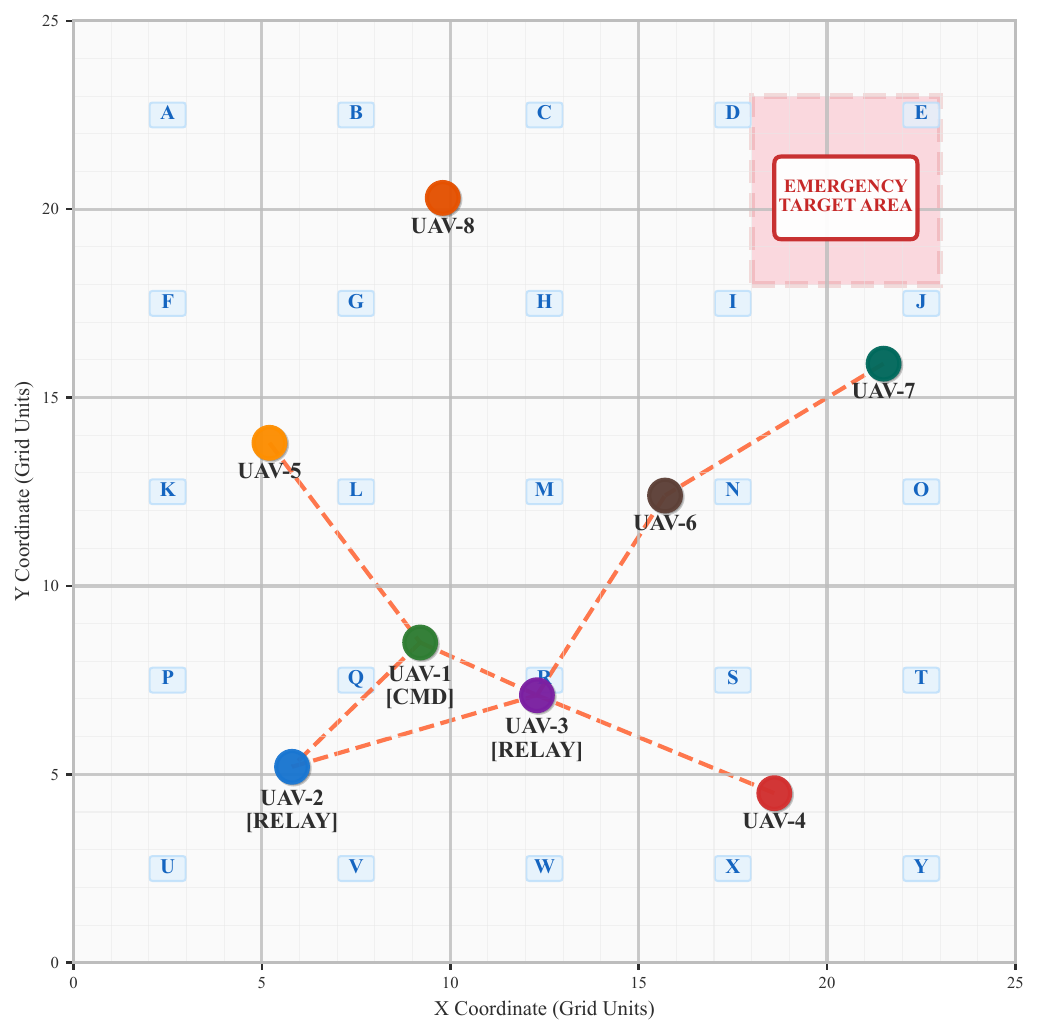}
    \caption{2D visualization of the Extreme scenario with hierarchical UAV coordination in a 25×25 grid map. The grid labels (A–Y) are provided solely for intuitive visualization and carry no actual operational meaning.}
    \label{fig:extreme_map}
\end{figure}

The extreme scenario is designed with 8 UAVs performing emergency tasks in a $25 \times 25$ grid environment, with its 2D visualization shown in Fig.~\ref{fig:extreme_map}. The system adopts a hierarchical command and control structure, where UAV-1 serves as the command node, and UAV-2 and UAV-3 act as relay nodes. All UAVs are randomly distributed across the map at initialization, and UAV-1 to UAV-3 are configured with communication distance constraints during the initial phase. The target task region is located in the upper-right corner of the map, with coordinate bounds $[18,\,23] \times [18,\,23]$, forming a $5 \times 5$ grid area.

To align with real-world UAV scenarios, we refer to the typical task coverage range of small- to medium-scale UAV swarms and set the grid scale such that one unit corresponds to an actual distance of 100 meters. This allows a $25 \times 25$ grid to cover a spatial area of $2.5\,\mathrm{km} \times 2.5\,\mathrm{km}$, with the UAV's movement speed set to $20\,\mathrm{m/s}$. Accordingly, each UAV can move $0.2$ grid units per second in the simulation environment. For ease of control and scheduling, the time step is set to 5 seconds, meaning that the UAV moves one grid unit per step.

The pipeline for autonomous semantic compression, communication, and task execution is shown in Algorithm~\ref{alg:1}. At the initial stage, the system sends the complete task prompt $M_{\text{raw}}$ only to the command node $u_1$, which contains the grid map structure, target region coordinates, velocity parameters, and the global initial positions ${\mathbf{p}_i^0}_{i=1}^{|\mathcal{U}|}$. Upon receiving the message, $u_1$ converts it into a compressed format $M_{\text{zip}}$ and forwards it to the relay node set $\mathcal{R} = \{u_2, u_3\}$. All other UAVs receive only system-level prompt instructions that guide them to parse and process $M_{\text{zip}}$, ensuring that each $u_i \in \mathcal{U}$ possesses LLM-based semantic understanding capability.

\begin{algorithm}[!htbp]
\SetAlgoLined
\caption{Pipeline for Autonomous Semantic Compression, Communication, and Task Execution}
\label{alg:1}
\KwIn{
UAV set $\mathcal{U}=\{u_1,\dots,u_N\}$; relay set $\mathcal{R}=\{u_2,u_3\}$; \\
initial positions $\{\mathbf{p}_i^0\}$; Task $M_{\text{raw}}$; $r_c$; $k_{\max}$; \\
$T_{\max}$; $S_{\max}$; $\Delta t$ \tcp*[f]{Environment parameters (map size, target region, speed, step length) defined in simulation setup}
}
\KwOut{Each $u_i$ either receives $M_{\text{zip}}$ or stays in standby}

Ensure $\text{dist}(u_p,u_q) < r_c$ for $p,q \in \{1,2,3\}$\;
$M_{\text{zip}} \gets \text{Compress}(M_{\text{raw}})$ by $u_1$\;
$\delta_1 \gets 1$, $\delta_{i\neq 1} \gets 0$\;

\For{$t=1$ \KwTo $S_{\max}$}{
  \uIf{$t=1$}{ 
    $u_1$ sends $M_{\text{zip}}$ to $\mathcal{R}$; mark $\delta=1$
  }
  \uElseIf{$t=2$}{ 
    Each $u_j \in \mathcal{R}$ sends to up to $k_{\max}$ neighbors within $r_c$ and $\delta=0$; mark $\delta=1$
  }
  \Else{ 
    Move each $u_i$ with $\delta_i=1$ one step\;
    Send to up to $k_{\max}$ new neighbors within $r_c$; mark $\delta=1$
  }
  \If{$t\cdot \Delta t \geq T_{\max}$}{\textbf{break}}
}

For $u_i$ with $\delta_i=0$: set standby mode\;
\end{algorithm}

The communication propagation follows a three-timestep hierarchical protocol. At $t = 1$, $u_1$ performs global perception and generates $M_{\text{zip}}$. At $t = 2$, $u_2$ and $u_3$ perceive the positions of nearby UAVs and transmit the message to UAVs satisfying $\text{dist}(u_i, u_j) < r_c$ and $\delta_j = 0$. Starting from $t = 3$, other UAVs that have received $M_{\text{zip}}$ begin task execution and rebroadcast the compressed message before each movement. The propagation remains constrained by a communication radius $r_c = 7.0$ and a maximum connection count $k_{\max} = 4$. Each UAV sets its reception status $\delta_i = 1$ only upon first receipt, to prevent redundancy and cyclic transmissions. An elimination mechanism is employed for handling communication failures: any UAV that fails to establish a connection with any $\delta_j = 1$ UAV throughout the execution remains in standby mode until the task ends. To ensure task feasibility, a maximum time limit $T_{\max} = 200$ seconds and a maximum movement step limit $S_{\max} = 40$ apply. The test is terminated once either condition is met.

\subsection{Design of System and Instruction Prompts}

\begin{myjsonbox}[label=box:instruction-json]{System Prompt JSON Example}
{
  "system_prompt": {
    "map": "25x25 grid",
    "current_time_step": 12,
    "uav_positions": {
      "UAV-1": {"role": "commander", "position": [5, 8]},
      "UAV-2": {"role": "relay", "position": [9, 6]},
      "UAV-3": {"role": "relay", "position": [7, 12]},
      "UAV-4": {"role": "executor", "position": [3, 21]},
      "...": "..."
    },
    "constraints": "..."
  }
}
\end{myjsonbox}

\begin{myjsonbox}[label=box:system]{Instruction Prompt JSON Example}
{
  "instruction_prompt": {
    "UAV-1": "You are UAV-1...commander. ...Emergency task: grid [18,23]x[18,23]... Send compressed instruction to UAV-2 and UAV-3.",
    "UAV-2": "You are UAV-2...relay node. ...Receive from UAV-1 and forward to nearby UAVs. Then move to target area.",
    "UAV-3": "...",
    "UAV-4": "You are UAV-4...executor. ...Receive compressed command and execute rescue action.",
    "...": "..."
  }
}
\end{myjsonbox}

The system prompt adopts a static template structure, where the current environmental information is dynamically filled in at each time step and continuously injected into the LLM context to support instruction compression and semantic understanding. This prompt is identical for all UAVs, and its structure is shown in the \texttt{System Prompt} code block. 

The task instruction prompt is injected once at the initialization stage of the task, does not change over time, and is designed in a differentiated manner according to the roles of different UAVs. In the extreme scenario, the system defines three types of roles: commander, relay, and executor, with their respective instruction templates shown in the \texttt{Instruction Prompt} code block.


\section{Experiments}

\subsection{Experimental Setup}

We conducted systematic experiments in four simulated environments. To comprehensively evaluate the proposed semantic compression communication mechanism for UAV swarms, we designed four evaluation metrics: Compression Ratio (CR), Semantic Preservation (SP), Bandwidth Utilization (BU), and Success Rate (SR).

\textbf{CR} measures the compression efficiency of the LLM-generated outputs. It is calculated based on the byte-length ratio of the instruction before and after compression, defined as follows. A lower CR indicates stronger compression ability and lower communication load.
\[
\text{CR} = \frac{\text{Bytes}_{\text{compressed}}}{\text{Bytes}_{\text{original}}}
\]

\textbf{SP} evaluates the fidelity of the compressed instructions in preserving the original task intent at the semantic level. We adopt the \texttt{DeBERTa-v3-xlarge-mnli} model~\cite{he2021debertav3} as a pretrained semantic alignment evaluator and compute the sentence-level similarity score between the compressed text and the original instruction using BERTScore. A higher score indicates better semantic preservation.

\textbf{BU} measures the communication efficiency of the system under limited bandwidth conditions. The calculation is as follows:
\[
\text{BU} = \frac{\sum_i \text{size}(M_i)}{B \cdot T}
\]
where $\text{size}(M_i)$ denotes the transmission size (in bits) of the $i$-th compressed message, $B$ represents the upper bound of the link bandwidth (in bits per second), and $T$ is the observation duration (in seconds). We set $B=1\times10^6$ bps (i.e., 1 Mbps) to simulate the typical upper bound of UAV link speed, and $T=60$ seconds to represent the communication window of a complete task phase.

\textbf{SR} reflects the system's ability to accomplish the task objectives. It is calculated using a simple ratio model:
\[
\text{SR} = \frac{N_{\text{reach}}}{N_{\text{total}}}
\]
where $N_{\text{reach}}$ is the number of UAVs that successfully reach the target area, and $N_{\text{total}}$ is the total number of UAVs participating in the task. To reflect the overall performance, we calculate the success rate under each of the four scenarios and take the average as the global evaluation metric.

\subsection{Overall Performance}

To comprehensively evaluate the performance of different LLMs in multi-UAV communication compression tasks, we selected 4 open-source models and 5 closed-source models. Due to the randomness in the initial positions of UAVs, each configuration was executed 10 times to ensure the stability and fairness of the results. We report the mean and standard deviation of the results, as shown in Table~\ref{tab:uav_compression}. Notably, the \texttt{LLaMA-4-Maverick-17B-128e-Instruct} model activates approximately 17B parameters during each inference, with a total parameter scale of around 400B~\cite{meta2025llama4-maverick-official}.

\begin{table*}[t]
  \centering
  \small
  \caption{Performance comparison of LLMs on the semantic compression communication for UAV swarms.}
  \label{tab:uav_compression}
  \begin{tabular}{lcccc}
    \toprule
    \textbf{Model} & \textbf{CR $\downarrow$} & \textbf{SP $\uparrow$} & \textbf{BU ($\times 10^{-5}$) $\uparrow$} & \textbf{SR $\uparrow$} \\
    \midrule
    \rowcolor{gray!15} \multicolumn{5}{l}{$\blacktriangledown$ \emph{Closed-source LLMs}} \\
    GPT-o3~\cite{openai2024o3} & 0.443 ± 0.028 & 0.729 ± 0.031 & 0.467 ± 0.035 & 0.932 ± 0.025 \\
    Claude-Sonnet-4-Thinking (20250514)~\cite{anthropic2025claude4} & 0.354 ± 0.022 & 0.493 ± 0.038 & 0.373 ± 0.028 & 0.586 ± 0.042 \\
    GPT-4o~\cite{hurst2024gpt} & 0.557 ± 0.033 & 0.813 ± 0.027 & 0.587 ± 0.041 & 0.953 ± 0.018 \\
    Claude-Sonnet-4 (20250514)~\cite{anthropic2025claude4} & 0.430 ± 0.026 & 0.728 ± 0.030 & 0.453 ± 0.034 & 0.932 ± 0.024 \\
    Grok-4~\cite{xaigpt2025grok4} & 0.709 ± 0.045 & 0.851 ± 0.023 & 0.747 ± 0.052 & 0.963 ± 0.016 \\
    \midrule
    \rowcolor{gray!15} \multicolumn{5}{l}{$\blacktriangledown$ \emph{Open-source LLMs}} \\
    DeepSeek‑V3-671B~\cite{liu2024deepseek} & 0.354 ± 0.021 & 0.751 ± 0.029 & 0.373 ± 0.029 & 0.938 ± 0.022 \\
    DeepSeek‑R1-671B~\cite{guo2025deepseek} & 0.354 ± 0.021 & 0.743 ± 0.031 & 0.373 ± 0.029 & 0.936 ± 0.023 \\
    Qwen3‑235B‑A22B~\cite{yang2025qwen3} & 0.684 ± 0.041 & 0.795 ± 0.026 & 0.720 ± 0.048 & 0.949 ± 0.019 \\
    LLaMA-4-Maverick-17B-128e-Instruct~\cite{meta2025llama4-maverick-official} & 0.241 ± 0.015 & 0.713 ± 0.037 & 0.253 ± 0.019 & 0.641 ± 0.048 \\
    \bottomrule
  \end{tabular}
\end{table*}

From the overall trend, most LLMs demonstrate high task success rates in the semantic compression task, indicating that the compressed instructions they generate maintain high transmission accuracy and executability. Moreover, there is a strong positive correlation between SP and SR, suggesting that the semantic fidelity of the compressed content is a key factor for successful task completion. Notably, \texttt{LLaMA-4-Maverick-128e-Instruct} achieves the lowest compression ratio, indicating strong compression capability; however, its task success rate is only 0.641, significantly lower than that of other models. Given its relatively low SP score, it can be inferred that the model sacrifices critical information in pursuit of extreme compression, making it difficult for execution nodes to accurately reconstruct the original task intent, thereby impairing overall performance as for reasoning-enhanced models, \texttt{GPT-o3}, \texttt{Claude-Sonnet-4-Thinking}, and \texttt{DeepSeek-R1} perform comparably to standard models, possibly indicating that reasoning capabilities have not yet yielded a distinct advantage in this specific task.

\subsection{Ablation Studies}

To evaluate the adaptability and stability of the proposed method under different task environments and group configurations, we designed ablation experiments based on two key variables: environmental complexity and cluster size, using \texttt{GPT-4o}—which shows stable performance across multiple metrics—as the unified model baseline.

\begin{figure}[htbp]
    \centering
    \includegraphics[width=0.9\linewidth]{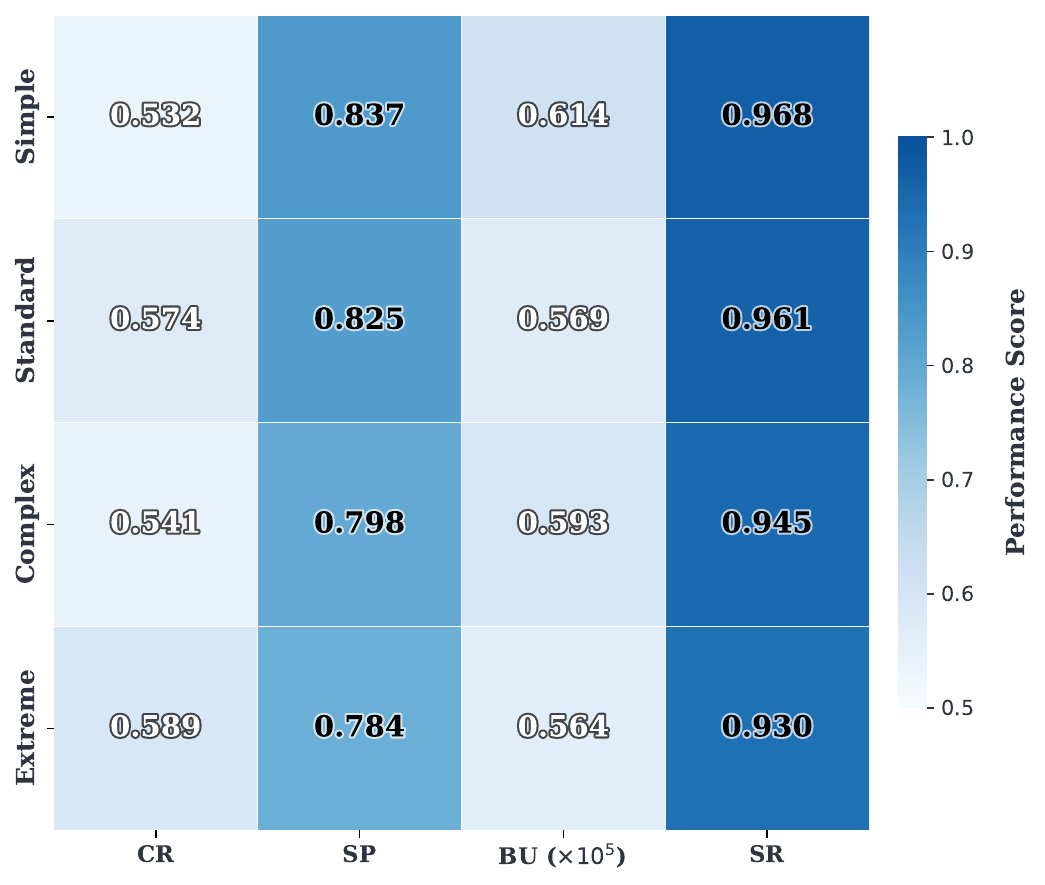}
    \caption{Heatmap of Performance Metrics across Environment Complexity Levels.}
    \label{fig:Heatmap}
\end{figure}

First, we analyzed one randomly selected execution result from ten repeated runs under four levels of environmental complexity. Fig.~\ref{fig:Heatmap} shows the heatmap distribution of the four core metrics across different environments. As the environment becomes more complex, both SP and SR exhibit a steady downward trend, indicating that task instructions are more susceptible to information loss caused by compression in complex scenarios. Meanwhile, CR and BU reach relatively high values in the \texttt{Extreme} environment, suggesting that the model may adopt more aggressive compression strategies under high-complexity conditions to cope with communication load constraints.

\begin{figure}[htbp]
    \centering
    \includegraphics[width=\linewidth]{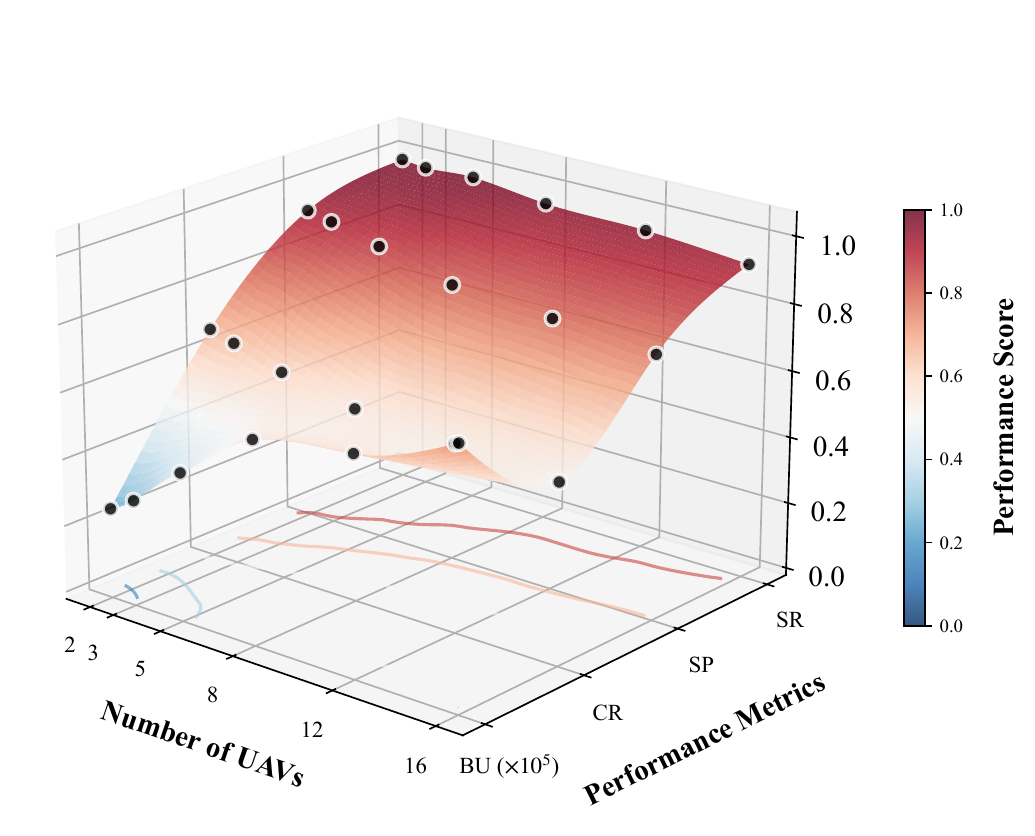}
    \caption{3D Surface Analysis of UAV Swarm Size under Extreme Conditions.}
    \label{fig:3D Surface}
\end{figure}

Next, to assess the impact of cluster size on system performance, we conducted ablation experiments under the \texttt{Extreme} environment with the number of UAVs ranging from 2 to 16. Fig.~\ref{fig:3D Surface} illustrates the 3D surface trend of the four metrics concerning varying group sizes. Overall, as the number of UAVs increases, SP and SR show a significant decline, indicating that higher interaction density within the cluster poses challenges to the consistency of compressed instruction execution. At the same time, CR gradually decreases while BU continuously increases, demonstrating that in larger-scale communication systems, the model tends to reduce compression intensity to preserve semantic transmission, thereby increasing bandwidth usage but improving interaction reliability.

\section{Conclusion}

This paper investigates the feasibility of an autonomous semantic compression communication mechanism for LLM-driven UAV swarms under constrained communication conditions, addressing the bandwidth bottleneck problem and verifying its potential to reduce communication load while preserving task semantic integrity. Future research directions include: first, extending the current 2D simulation environment to high-fidelity 3D scenarios to more realistically simulate spatial environments and communication link characteristics; second, further exploring multimodal semantic compression methods for MLLM-based UAV swarms to achieve efficient transmission and fusion of cross-modal task information, thereby comprehensively enhancing swarm collaborative communication and decision-making capabilities in complex task environments.

\bibliographystyle{IEEEtran}
\bibliography{egbib}

\end{document}